\documentclass[conference]{IEEEtran}
\IEEEoverridecommandlockouts
\usepackage{cite}
\usepackage{amsmath,amssymb,amsfonts}
\usepackage{algorithmic}
\usepackage{graphicx}
\usepackage{textcomp}
\def\BibTeX{{\rm B\kern-.05em{\sc i\kern-.025em b}\kern-.08em
    T\kern-.1667em\lower.7ex\hbox{E}\kern-.125emX}}
\usepackage{diagbox}
\usepackage{arydshln}
\usepackage{import}
\usepackage{pgfplots}
\pgfplotsset{compat=1.13} 

\newcommand{\myheader}{Accepted as a conference paper at the \textit{International Joint Conference on Neural Networks (IJCNN) 2018,  \textcopyright\ 2018 IEEE}}

\usepackage{atbegshi,picture}

\usepackage{datetime}            % used to put a date in the header

\usepackage{lastpage}            % organises page numbering

\newcommand{\myleftstd}{1.6in}

\ifthenelse{\isodd{\thepage}}
 {\newcommand{\myleftmargin}{\oddsidemargin+\myleftstd}}
 {\newcommand{\myleftmargin}{\evensidemargin+\myleftstd}}

\AtBeginShipoutNext { 
 \AtBeginShipoutUpperLeft{%
  \put(\dimexpr 0.2\myleftmargin\relax,-1.2cm){\parbox{\textwidth}{\footnotesize\sffamily\noindent{\centering \myheader \hfill}}}%
}}
\begin{document}

\title{Image Generation and Translation\\with Disentangled Representations\thanks{The authors gratefully acknowledge partial support from the DFG under project CML (TRR 169) and the EU under project SECURE (No 642667).}
}

\author{\IEEEauthorblockN{Tobias Hinz}
\IEEEauthorblockA{Department of Informatics \\
Universit\"at Hamburg\\
Hamburg, Germany \\
hinz@informatik.uni-hamburg.de}
\and
\IEEEauthorblockN{Stefan Wermter}
\IEEEauthorblockA{Department of Informatics \\
Universit\"at Hamburg\\
Hamburg, Germany \\
wermter@informatik.uni-hamburg.de}
}

\maketitle

\begin{abstract}
Generative models have made significant progress in the tasks of modeling complex data distributions such as natural images. The introduction of Generative Adversarial Networks (GANs) and auto-encoders lead to the possibility of training on big data sets in an unsupervised manner. However, for many generative models it is not possible to specify what kind of image should be generated and it is not possible to translate existing images into new images of similar domains. Furthermore, models that can perform image-to-image translation often need distinct models for each domain, making it hard to scale these systems to multiple domain image-to-image translation.

We introduce a model that can do both, controllable image generation and image-to-image translation between multiple domains. We split our image representation into two parts encoding unstructured and structured information respectively. The latter is designed in a disentangled manner, so that different parts encode different image characteristics. We train an encoder to encode images into these representations and use a small amount of labeled data to specify what kind of information should be encoded in the disentangled part. A generator is trained to generate images from these representations using the characteristics provided by the disentangled part of the representation. Through this we can control what kind of images the generator generates, translate images between different domains, and even learn unknown data-generating factors while only using one single model.
\end{abstract}


\section{Introduction}
The introduction of Generative Adversarial Networks \cite{goodfellow2014generative} (GANs) provided a way to generate realistic images through a model that can be trained in an unsupervised fashion. While it has been observed that images produced by GANs can be sharp and realistic, the original GAN model does not provide any control over what kind of image is generated. Furthermore, it does not provide a way to modify existing data samples, but can only generate new ones. Since then, GANs have been extended to also support or handle tasks such as image-to-image translation and controllable image generation, two tasks that require the modeling of high-dimensional data and a certain amount of understanding about the content of images.

Image-to-image translation takes as input some image and tries to ``translate'' it into a different domain. This can, for example, include changing the overall style of the image \cite{Isola_2017_CVPR}, translating the objects within the image into similar ones \cite{Zhu_2017_ICCV, liu2017unsupervised} or manipulating certain aspects of the image, e.g.\ by changing facial characteristics \cite{lample2017fader, shen2017learning}. One difficulty in image-to-image translation is that it is often an unsupervised problem, i.e.\ we do not have a ground truth of what the translated image should look like. If we have, for example, the image of a male face and want to translate it into a female face, we usually have no image to compare it with and there are many different ways in which a male face could be modified to look more like a female one. Additionally, many current techniques need to train individual translators for each domain.
% i.e.\ they need one translator to translate a smiling face into a non-smiling face, one translator to translate a male face into a female face, etc.
This quickly becomes unfeasible as the number of domains increases, since for $k$ domains $k(k-1)$ translators would be needed.

Controllable image generation is a related problem in which we want to exert some control over what kind of image is generated. This could for example mean specifying what kind of a digit is generated or whether a generated face should be male or female \cite{mirza2014conditional, spurr2017ecml, zhang2017structured}. This is usually achieved by providing a label to the generator and discriminator. Since the discriminator gets correctly labeled data samples from the real data distribution it learns to associate the labels with specific features in the images. In order to fool the discriminator the generator then learns to generate images that correspond to the provided labels. While this requires a (partially) labeled training set, it provides us with more control over what kind of images are generated and has also been shown to improve the image quality \cite{salimans2016improved}.

So far, many of the methodologies focus on either image generation or image translation, but can rarely do both tasks. However, working in the domain of images there is conceptually not a big difference between translating images from one domain into another, or generating a new image according to certain conditions. Furthermore, many of the systems that perform image translation need distinct translators for individual domains -- an approach that does not scale well with multiple domains. Additionally, many of the approaches need labeled training images for each of the domains they work with. Finally, most image translation methods encode the image information in entangled representations without easy access to the domain information or sometimes even exclude the domain information entirely from the image representation. 

Our approach, on the other hand, aims at performing both, generating new images and translating between multiple domains with only one model. It does only need few labeled training samples and encodes all information into the representation for the generator to use. Information important for the respective domains is encoded in a disentangled manner and we can even detect unknown data-generating factors without the need for any labels. For this, we make use of a generator $G$ that generates an image $X$ from a vector $Z$, and an encoder $E$ that encodes images $X$ into a representation $Z$. In order to gain control over the images, $Z$ is divided into two parts $(u, c)$. Here, $u$ encodes image characteristics that we do not want to model explicitly, while $c$ encodes characteristics that we want to control, e.g. which digit of the MNIST data set should be generated.

We then train the encoder to encode provided labels into $c$ and all other information into $u$, while the generator is trained to construct realistic images from $(u,c)$.
%For this, we use the reconstruction loss between the original image $X$ and the reconstructed image $X' = G(E(X))$ and maximize the mutual information between $c$ and the image generated from it.
This methodology already offers the possibility of translating images by using the encoder to get an image representation $Z = (u, c)$, changing $c$ to the desired domain (e.g.\ changing the image class) and using the generator to generate the translated image. 
% In order to control the kind of images that are generated by the generator $G$ we need to ensure that $G$ uses the information provided in $c$. We achieve this by forcing $G$ to generate images from which we can reconstruct $c$ by maximizing the mutual information $I$ between $c$ and $G(u,c)$.
%However, it does not offer a way to generate new data samples, since we do not know the distribution of $u$. To remedy this 
To also offer a way to generate new data samples we introduce a discriminator $D$ that takes as input the representation $Z$ and the according image $X$ and tries to determine if the pair $(Z, X)$ came from the generator or the encoder. 
%The inputs to the discriminator are either real training samples $X$ and the representation $Z' = E(X)$ or a randomly sampled representation $Z$, where $u$ is sampled from a uniform distribution, and the generated image $X' = G(Z)$. 
Both the encoder and the generator try to build pairs $(X, Z)$ that are classified incorrectly by the discriminator. This has two beneficial effects: firstly, it encourages the encoder and the generator to learn inverse functions of each other \cite{dumoulin2017adversarially} which should minimize the reconstruction error $X - G(E(X))$, and secondly, it forces the generator to generate meaningful and controllable images from randomly sampled $Z$.

To summarize: we propose a model that can do both controllable image generation and image-to-image translation between multiple domains based on information encoded within the representation. We only need very few labeled training examples (less than 2\% on all tested data sets) and can even detect unknown data-generating factors, which can also be used for controllable image generation and translation. All information is directly encoded in the latent representation in a disentangled manner to which both the generator and the encoder have unrestricted access at all times during training.

\begin{figure*}
\centering
\includegraphics{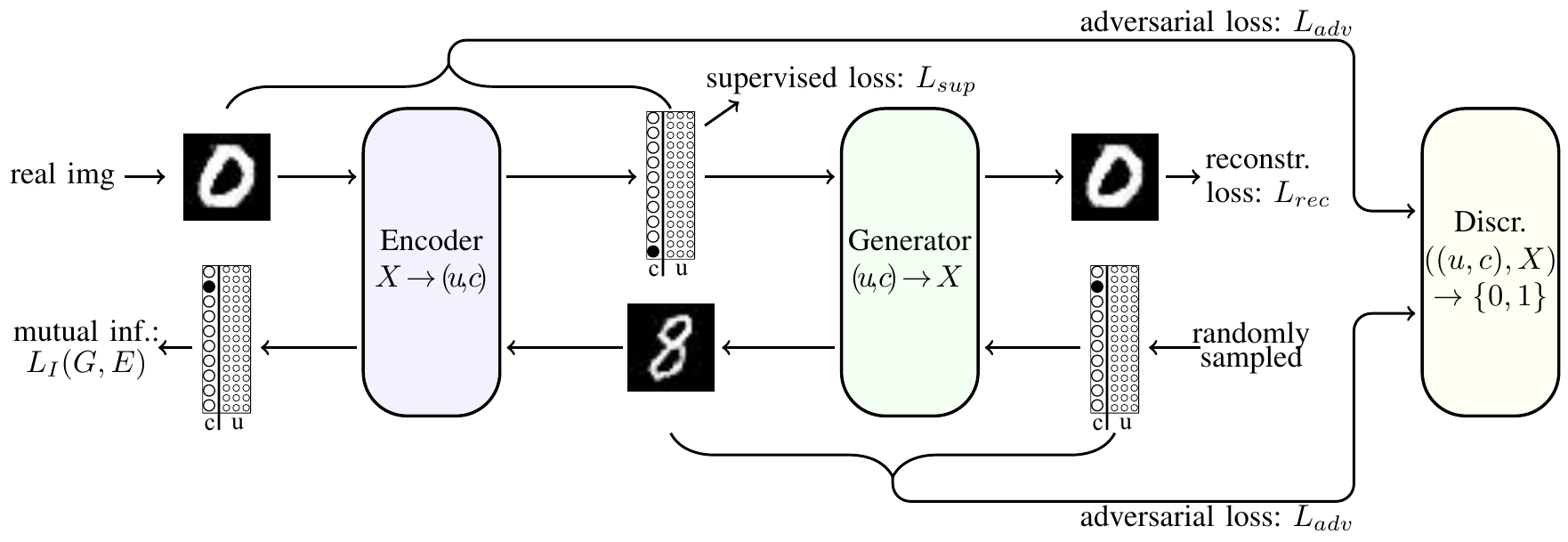}
\caption{The encoder gets as input an image from the training set and possibly the associated label. It then encodes the image into a latent representation which is used by the generator to reconstruct the image. The generator gets as input a latent representation (either from the encoder or randomly sampled) and generates an image from it, while the encoder tries to reconstruct $c$ from the image. The discriminator tries to determine whether pairs of latent representation and image come from the encoder or the generator.}
\label{fig:model_architecture}
\end{figure*}

\section{Related Work}
Mirza et al. \cite{mirza2014conditional} introduced conditional GANs by supplying both the generator and the discriminator with labels. Perarnau et al. \cite{perarnau2016invertible} train two encoders with the help of a previously trained conditional generator where one encoder maps the image into a representation while the second encoder maps the image into a condition (e.g.\ the class label). Using the learned representation of an image a new condition can be specified to translate the original image. However, this approach only trains the encoder after the generator has already been trained, limiting the possibilities of interaction between the generator and the encoders. Additionally, all real data that is used during training the encoders needs to be labeled. Both Donahue et al. \cite{donahue2017adversarial} and Dumoulin et al. \cite{dumoulin2017adversarially} suggest training the encoder jointly at the same time as the generator. While this offers ways of encoding images, it does not offer any controllability over what kind of images are generated, nor does it offer the possibility to translate images into other domains.

Chen et al. \cite{chen2016infogan} introduced GANs for disentangled representations. Training proceeds completely unsupervised and the model learns to disentangle underlying data-generating factors, which are modeled directly in the latent representation. These disentangled representations can then be used to control what kinds of images are generated. However, there is no control over which data-generating factors are learned and they do not necessarily coincide with human-interpretable factors (e.g.\ class labels). Spurr et al. \cite{spurr2017ecml} extended this setting by incorporating some labels into the training process. Through this they have more control over which data-generating factors the generator learns, while still being able to also learn unlabeled or unknown data-generating factors. In addition to the generator and discriminator Li et al. \cite{li2017triple} use a classifier to achieve a controllable generator. The data distribution is characterized by the classifier and the generator, while the discriminator only focuses on distinguishing between real and generated samples. However, all three systems lack the capability of mapping existing images into latent representations, i.e.\ they do not offer the possibility of image-to-image translation. Hinz et al. \cite{hinz2018inferencing} introduced a model that learns disentangled representations for both generated and real data. While this model learns representations for existing data it can not translate between different domains and there is no control over which kind of data-generating factors are learned.

Zhang et al. \cite{zhang2017structured} developed a model that is capable of both generating new images and translating existing ones. For this they split up the latent representations into two parts, encoding label information and other (unstructured) information. The labels are used for semi-supervised training and two additional inference networks are introduced which map images into the two parts of the representation. However, the model can only translate images based on the labels that are originally supplied, i.e.\ it cannot learn new data-generating factors and it needs individual inference networks for different parts of the representation. Choi et al. \cite{choi2017stargan} can also perform multi-domain image translation within one model, but require a fully labeled training set or the need to employ a specifically developed technique to deal with samples that are not fully labeled. 

Shen et al. \cite{shen2017learning} model the image manipulation operation as learning the residual, i.e.\ learning the ``difference image'' between the original and the translated image. For this, they need two networks for each attribute in order to model inverse operations, e.g.\ one network to add eyeglasses and one network to remove glasses. Lample et al. \cite{lample2017fader} use an encoder-decoder architecture for manipulating certain attributes of faces while keeping the rest of the image constant. This is achieved by making the representations invariant to the attributes, i.e.\ a representation encodes a face without the learned attributes and the generator gets the ``base'' representation plus the desired attributes. They need labels for each image during training and can also not detect novel data-generating factors. The procedure also imposes constraints on the representation, since it cannot encode the respective attributes which might have negative impacts on the quality of the learned representations.

\section{Methodology}
Our model consists of a generator $G$, an encoder $E$, and a discriminator $D$. The generator takes as input a vector $Z$ and transforms it into an image $X$, while the encoder takes as input an image $X$ and maps it into a latent representation $Z$. The discriminator takes as input both an image $X$ and a latent representation $Z$, and tries to determine whether the pair came from the generator or the encoder. Fig.~\ref{fig:model_architecture} gives a high-level overview over our model. The latent representation $Z$ is split up into two parts $(u, c)$, where $u$ encodes unstructured information and noise, while $c$ encodes structured information and data-generating factors such as for example class labels.

$G$ and $E$ work together in that $E$ takes as input an image, maps it into a latent representation, and $G$ uses this representation to generate an image from it. Alternatively, the generator gets as input a randomly sampled representation $Z$ and transforms it into an image from which the encoder tries to infer $c$. This is to ensure that the generator indeed uses the information provided in $c$ as well as to detect previously unknown data-generating factors. The discriminator and its adversarial loss are used to improve the image quality and to encourage $G$ and $E$ to model inverse functions.

In order to ensure that the latent representation $Z$ encodes the information that is needed to reconstruct the original image, we minimize the reconstruction loss, where $E$ is the encoding part and $G$ takes the role of the decoder:
\[
\underset{G, E}{\text{min}}\ L_{rec}(G, E) = \mathbb{E}_{X\sim P_{\text{data}}}[\vert\vert X - G(E(X))\vert\vert_2^2].
\]

In contrast to other approaches, we do not condition the generator on additional labels to control the image generation process, but instead encode all the necessary information directly within the latent representation $Z$, more specifically in $c$. To achieve this, $c$ is made up of both categorical values $c_{cat}$ and continuous values $c_{cont}$ and the generator learns to associate these values with certain attributes or characteristics. To achieve this, we maximize the mutual information $I(c, G(u,c))$, i.e.\ the mutual information between $c$ and the images generated from $(u, c)$. Maximizing the mutual information directly is hard as it requires the posterior $P(c\vert x)$, and we therefore follow the approach by Chen et al. \cite{chen2016infogan} and define an auxiliary distribution $E(c\vert x)$ to approximate $P(c\vert x)$, where $E$ is parameterized by our encoder. We then maximize the lower bound 
\begin{equation*}
\begin{split}
\underset{G, E}{\text{max}}\ L_I(G, E) = \mathbb{E}_{c\sim P(c), u\sim P(u), X\sim G(u,c)}[log\ E(c\vert X)]
\\
+H(c) \leq I(c;G(u,c)).
\end{split}
\end{equation*}
For simplicity reasons, we fix the distribution over $c$ and, therefore, the entropy term $H(c)$ is treated as a constant.

While this approach is completely unsupervised and can detect meaningful characteristics that are encoded through $c$ by itself, it has the drawback that we cannot specify certain characteristics that we want to be encoded within $c$ (e.g.\ class labels). To remedy this, we introduce an additional cost, i.e.\ for a small subset of labeled data points we train $E$ in a supervised manner to encode the provided information within $c$:
\[
\underset{E}{\text{min}}\ L_{sup}(E)=-\sum_i y_i\ log(y_i^\ast)
\]
where $y_i$ are labels such as class labels or characteristics like hair color or the presence of glasses and $y_i^\ast$ are the encoder's predictions. Crucially, this process is only used to ``guide'' the encoder into associating certain specified information with given parts of $c$. Since the generator and the encoder are trained in a joint fashion (for maximizing the mutual information $I(c, G(u, c))$ and the reconstruction loss), the generator quickly picks up on the characteristics and their associated encodings and generates images with similar characteristics for similar encodings.

As a result, we only need comparatively few labeled data points for this process and the data points need not even be fully labeled. For example, if we have an image of a person's face with a given label that indicates the presence or absence of a smile, we can use this label to train the encoder to associate the presence or absence of a smile with a given part of $c$ while ignoring the parts of $c$ that encode other information. This also means that we can still use the model to learn to encode unlabeled characteristics within $c$ while using other parts of $c$ to encode predetermined characteristics.

Finally, to ensure that the images generated by $G$ are realistic, that $E$ and $G$ learn to model inverse functions, and that $E$ models $u$ according to the chosen distribution we use a discriminator $D$. This discriminator tries to determine whether a pair of a latent representation and the associated image, i.e.\ either $((u, c), G(u, c))$ or $(X, E(X))$, comes from the encoder $E$ or the generator $G$. Both $E$ and $G$ try to learn transformations that make the discriminator mis-classify their representation-image pair, i.e.\
\begin{equation*}
\begin{split}
\underset{G, E}{\text{min}}\ \underset{D}{\text{max}}\ L_{adv}(D, G, E) = \mathbb{E}_{X\sim P_{\text{data}}}[logD(X, E(X))]
\\
+ \mathbb{E}_{Z\sim P_Z}[log(1-D(G(Z), Z))].
\end{split}
\end{equation*}

This leads to our final objective function for training the whole model:
\begin{equation*}
\begin{split}
\underset{G, E}{\text{min}}\ \underset{D}{\text{max}}\ L(D, G, E) = \lambda_1 L_{sup} + \lambda_2 L_{rec} \\
 - \lambda_3 L_I + \lambda_4 L_{adv},
 \end{split}
\end{equation*}
where $\lambda_i, i=1...4$ are used to weight the impact of the individual loss terms. Given this model, we can now encode real or generated images into a latent representation $Z = (u, c)$. The characteristics learned in $c$ are either specified through a limited amount of labeled training data or discovered in an unsupervised way by $G$ and $E$. We can now use $G$ to generate new images with given characteristics specified by $c$. We can also use $G$ and $E$ for image-to-image translation by obtaining the representation of a given image through $E$, modifying image characteristics by changing values in $c$, and generating the new image with updated image characteristics.

\section{Implementation}
\setlength\tabcolsep{1.5pt} % default value: 6pt
\begin{table}
\centering
\begin{tabular}{| c | c |}
\hline
\textcolor{blue}{MNIST} & \textcolor{blue}{SVHN / CelebA} \\ 
\hline
\multicolumn{2}{| c |}{\textcolor{purple}{Discriminator}} \\
\multicolumn{2}{| c |}{dropout with probability 0.3 after each layer} \\
\hdashline
\multicolumn{2}{| c |}{image $X$} \\
\hdashline
3x3 conv. 64, BN, ELU, stride 2 & 4x4 conv. 64, BN, ELU, stride 2 \\
3x3 conv. 128, BN, ELU, stride 2 & 4x4 conv. 128, BN, ELU, stride 2 \\
& 4x4 conv. 256, BN, ELU, stride 2 \\
FC 512, BN, ELU & FC 1024, BN, ELU \\
\hdashline
\multicolumn{2}{| c |}{representation $Z$} \\
\hdashline
1x1 conv. 64, BN, ELU, stride 1 & 1x1 conv. 64, BN, ELU, stride 1 \\
1x1 conv. 128, BN, ELU, stride 1 & 1x1 conv. 128, BN, ELU, stride 1 \\
& 1x1 conv. 256, BN, ELU, stride 1 \\
FC 512, BN, ELU & FC 1024, BN, ELU \\
\hdashline
\multicolumn{2}{| c |}{concatenate across channel axis} \\
\hdashline
FC 1024, BN, ELU & FC 1024, BN, ELU \\
FC 1, Sigmoid & FC 1, Sigmoid \\
\hline

\multicolumn{2}{| c |}{\textcolor{purple}{Generator}} \\
FC 3136, BN, ELU & FC 2048, BN, ELU \\
reshape 7x7x64 & reshape 4x4x128 \\
4x4 deconv. 128, BN, ELU, stride 2 & 4x4 deconv. 128, BN, ELU, stride 2 \\
4x4 deconv. 64, BN, ELU, stride 1 & 4x4 deconv. 64, BN, ELU, stride 2 \\
 & 4x4 deconv. 32, BN, ELU, stride 2 \\
4x4 deconv. 1, Sigmoid, stride 2 & 3x3 deconv. 3, Sigmoid, stride 1 \\

\hline
\multicolumn{2}{| c |}{\textcolor{purple}{Encoder}} \\
3x3 conv. 32, BN, ELU, stride 1 & 3x3 conv. 32, BN, ELU, stride 1 \\
3x3 conv. 64, BN, ELU, stride 2 & 3x3 conv. 64, BN, ELU, stride 2 \\
3x3 conv. 128, BN, ELU, stride 2 & 3x3 conv. 128, BN, ELU, stride 1 \\
 & 3x3 conv. 256, BN, ELU, stride 2 \\
 & 3x3 conv. 512, BN, ELU, stride 2 \\
FC 1024, BN, ELU & FC 1024, BN, ELU  \\
FC output layer & FC output layer \\
\hline
\end{tabular}
\caption{Overview of our network architectures.}
\label{tab:network_architecture}
\vspace{-1em}
\end{table}
\setlength\tabcolsep{6pt} % default value: 6pt

The generator is implemented as a deconvolutional neural network, while the encoder and the discriminator are convolutional neural networks. For the SVHN and the CelebA data sets we use the same architecture, while we use slightly smaller networks for the MNIST data set. For an overview of the used architectures see Table~\ref{tab:network_architecture}.

In our experiments the weight $\lambda_1$ for the supervised loss was set to 10 to ensure that the encoder learns the labeled data correctly, the weighting $\lambda_2$ of the reconstruction loss was set to 1, and $\lambda_3$ and $\lambda_4$ were linearly increased from 0 to 1 during the first 1000 (10000) iterations on the MNIST (SVHN, CelebA) data set. We train the model for 50000, 150000 and 300000 iterations respectively on the MNIST, the SVHN, and the CelebA data set. The learning rate is 0.0001 for the discriminator and 0.0003 for the generator and the encoder, and the batch size is 64 in all experiments. For training, we use the Adam optimizer \cite{kingma2015adam} with $\beta_1 = 0.5$ and $\beta_2 = 0.999$.

\begin{figure*}
\centering
\import{}{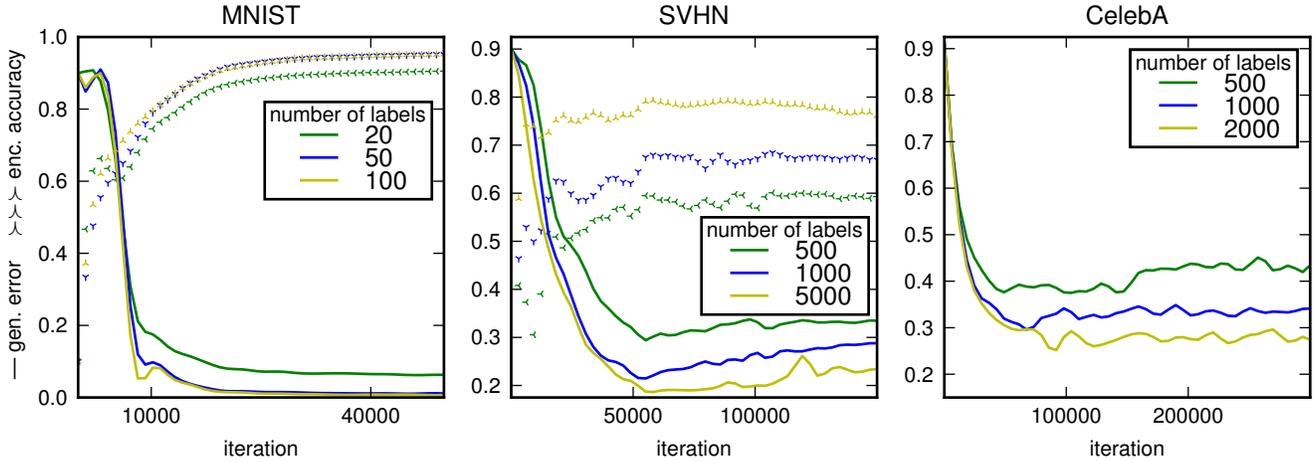}
\vspace{-1em}
\caption{Development of generator and encoder accuracy averaged over ten independent runs for the different data sets. The dotted lines represent the encoder accuracy, while the continuous lines depict the error of the generator, i.e.\ percentage of inaccurately generated images.}
\label{fig:generator_encoder_acc}
\end{figure*}

Since we only use a small amount of labeled data, we initially favor drawing labeled samples from the training set. In the beginning of the training process the probability of drawing labeled data is therefore one. During the first 1000 (10000) iterations on the MNIST (SVHN, CelebA) data set this probability is linearly decreased until it reaches the actual labeled sample ratio in the data set. For our latent representation, $u$ is sampled from a uniform distribution $U(-1,1)$, while $c$ is split up into categorical and continuous variables $c_{cat}$ and $c_{cont}$. For the categorical variables we use the softmax activation in the final layer, while the continuous variables are modeled as a factored Gaussian.

\section{Experiments}
We test our model on the MNIST, the SVHN, and the CelebA data set. On the MNIST and the SVHN data set the labeled information consists of class labels (digit type), while on the CelebA data set the labeled information contains attributes such as hair color and gender. For each data set we train a classifier to quantitatively assess the controllability of our generator and qualitatively examine the model's capability to translate images from the respective test sets.

\begin{figure}
\footnotesize
\centering
\def\svgwidth{.72\columnwidth}
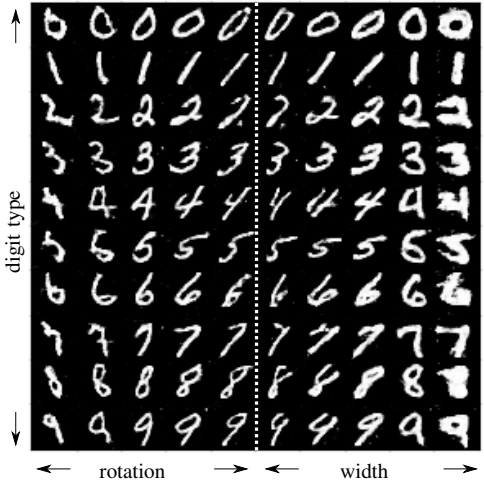
%\centering
%\includegraphics[scale=0.8]{./figures/mnist/mnist_random_samples.png}
\caption{Image generation (model trained with 100 labels): for each column the random vector $u$ is kept constant, while the categorical variable encoding class information is changed for each row. In the first five columns we vary $c_1$ from -1 to 1, in the second five columns we vary $c_2$ from -1 to 1.}
\label{fig:mnist_random_samples}
\end{figure}

On the MNIST data set we model $u$ as a 16-dimensional vector, while $c$ consists of a 10-dimensional categorical variable $c_{cat}$ which encodes class information and two continuous variables $c_1, c_2\sim U(-1,1)$. At regular intervals during the training process we generate 500 new images of each class by specifying $c_{cat}$ accordingly and use a previously trained classifier (99.43\% accuracy on the MNIST test set) to classify them. The development of the generator's accuracy (averaged over 10 independent runs) is depicted in Fig.~\ref{fig:generator_encoder_acc}, and Table~\ref{tab:generator_accuracy_mnist} gives the average accuracy at the end of training. We can see that for 50 and 100 samples the generator generates the desired digits with a very high accuracy and some generated samples are shown in Fig.~\ref{fig:mnist_random_samples}.

\begin{table}[b]
\centering
\setlength{\tabcolsep}{0.7em}
\begin{tabular}{| c | c c c |} \hline
\diagbox{Model}{num labels} & 20 & 50 & 100\\ \hline
Triple-GAN \cite{li2017triple} & 3.06 & 1.80 & 1.29 \\
SGAN \cite{zhang2017structured} & 1.68 & 1.23 & 0.93 \\
Ours & 6.29 $\pm$ 4.08 & 1.09 $\pm$ 1.09 & 0.66 $\pm$ 0.17 \\ \hline
\end{tabular}
\caption{Errors (\%) of generated samples on MNIST.}
\label{tab:generator_accuracy_mnist}
\vspace{-1.5em}
\end{table}

When we use only 20 labeled samples the generator sometimes ''mixes up`` two classes (e.g. 4 and 9), which then leads to a lower accuracy of only roughly 80\%. The chances of this happening are around 50\% and this explains the comparatively low generator accuracy when only 20 labels are used. This problem does not occur when marginally more samples are used and for 50 labeled samples this problem did not occur anymore. Zhang et al. \cite{zhang2017structured} achieve a good performance even with 20 labels but have to use two distinct inference networks, one for the class information and one for the unstructured part of the representation. Spurr et al. \cite{spurr2017ecml}, who only focus on learning a controllable generator, report that they need a minimum of 132 labels to achieve their goal.

Not only do we have control over what kind of digit is generated by the generator, but we can also control the stroke width and digit rotation by modifying $c_1$ and $c_2$. These characteristics were identified without any labels by maximizing the mutual information between $c_1$, $c_2$, and the images generated from them. Fig.~\ref{fig:mnist_random_samples} also shows these characteristics by interpolating $c_1$ from -1 to 1 in the first five columns and $c_2$ in the last five columns. 

Fig.~\ref{fig:generator_encoder_acc} also shows the development of the encoder's accuracy measured on the test set. We can see that the accuracy approaches around 96\% for 50 and 100 labels. While this is not the optimal known state-of-the art for this number of labels we still achieve reasonable performance even though this is not our main training criterion. Other approaches that outperform ours with the same amount of labels usually focus explicitly on the encoder accuracy and the system is trained to determine the digit classes. Our model, on the other hand, does not only encode the digit identity, but also other information about the image such as general style, stroke width, and digit rotation.

\begin{figure}
\centering
\includegraphics[scale=0.8]{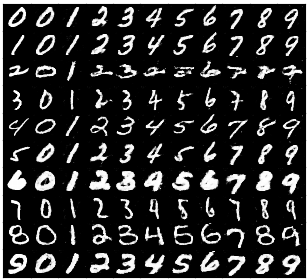}
\caption{Image translation: first column are randomly sampled images from the MNIST test set, other columns are translated images where only the categorical value of c was adapted.}
\label{fig:mnist_class_translation}
\end{figure}

\begin{figure}
\centering
\includegraphics[scale=0.8]{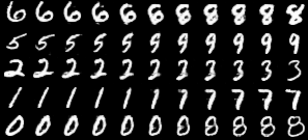}
\caption{Image interpolation: the first and last columns are images sampled from the MNIST test set, the intermediate columns are linear interpolations.}
\label{fig:mnist_interpolation}
\vspace{-1em}
\end{figure}

Finally, we show that we can translate images from the test set into other classes in Fig.~\ref{fig:mnist_class_translation} and interpolations between test set images in Fig.~\ref{fig:mnist_interpolation}. The first column in Fig.~\ref{fig:mnist_class_translation} depicts images from the MNIST test set, while the other columns show the translation of that image to other digit classes. We can see that general style information as well as stroke width and digit rotation are consistent across the translations. In Fig.~\ref{fig:mnist_interpolation}, the first and last columns show images from the MNIST test set, while the intermediate columns depict linear interpolations between the two images. We see that the interpolations progress smoothly and often change class identities around midway through the interpolation, when the ``class label'' in the representation has the same probability for the respective start and end classes.

On the SVHN data set we model $u$ as a 128-dimensional vector, while $c$ consists of four categorical variables and four continuous variables. We use $c_{cat_1}$ to encode the class information (10 classes), while the other three categorical variables are five-dimensional. Again, we track the performance of the generator during training with the help of a previously trained classifier (94.51\% accuracy on the test set). Fig.~\ref{fig:generator_encoder_acc} shows that the accuracy of both the generator and the encoder increases with the amount of labels, while Table~\ref{tab:generator_accuracy_svhn_celeba} shows the average performance of the generator after training is completed. We achieve a reasonably good performance with only 1000 labeled samples, as opposed to Spurr et al. \cite{spurr2017ecml} who need more than 7000 labeled samples to achieve a similar generator accuracy.

\begin{figure}
\centering
\includegraphics[scale=0.98]{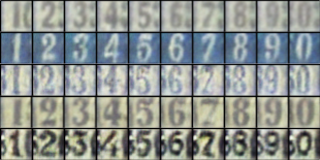}
\caption{Image generation (model trained with 1000 labels): the representation is kept constant for each row while the variable encoding class information changes across the columns.}
\label{fig:svhn_random_samples}
\vspace{-1em}
\end{figure}

\begin{figure}
\centering
\includegraphics[scale=.89]{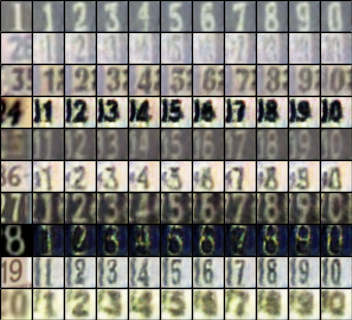}
\caption{Image translation: the first column contains randomly sampled images from the SVHN test set, while the other columns are translated images where the variable encoding class information is changed across columns.}
\label{fig:svhn_class_translation}
\end{figure}

\begin{figure}
\footnotesize
\centering
\def\svgwidth{\columnwidth}
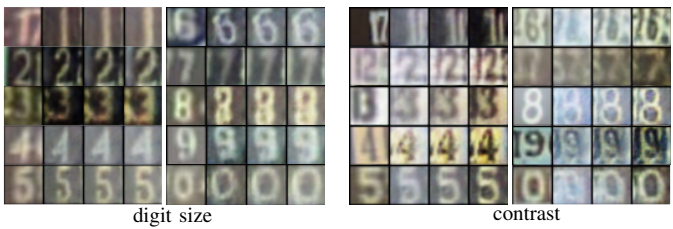
\caption{Image translation: the first column of each block shows randomly sampled images from the SVHN test set. The other columns are translated images where one of the continuous variables $c_{cont}$ is changed from -1 to 1.}
\label{fig:svhn_translations_cont}
\vspace{-1em}
\end{figure}

\begin{figure}
\centering
\includegraphics[scale=0.89]{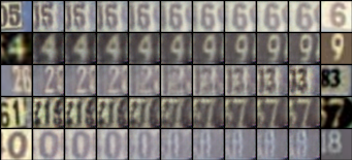}
\caption{Image interpolation: the first and last columns are images sampled from the SVHN test set, the intermediate columns are linear interpolations.}
\label{fig:svhn_interpolation}
\vspace{-1em}
\end{figure}

Fig.~\ref{fig:svhn_random_samples} shows samples of different classes generated by the generator. We can see that stylistic information is kept constant across the different classes. This indicates that $c_{cat_1}$ indeed captures the label information, while the rest of the representation $Z$ encodes other stylistic information. Fig.~\ref{fig:svhn_class_translation} shows translations of images from the SVHN test set. Again, we can see that the stylistic information is conserved across the different images, while the digit is controlled by $c_{cat_1}$. Fig.~\ref{fig:svhn_translations_cont} shows translations where only one of the continuous variables is changed across columns. Here, we show examples of two of the continuous variables, which learned to encode the digit size and the contrast. These factors were learned completely unsupervised and without any supplied labels. Finally, Fig.~\ref{fig:svhn_interpolation} shows interpolations between images from the test set. Again, the interpolations progress in a smooth manner and, as before, the digit identity changes around midway through the interpolation.

\begin{table}[b]
\centering
\setlength{\tabcolsep}{0.7em}
\begin{tabular}{| c | c c c | c c c |}
\hline
Data Set & \multicolumn{3}{c |}{SVHN} & \multicolumn{3}{c |}{CelebA} \\
\hline
Number of Labels & 500 & 1000 & 5000 & 500 & 1000 & 2000 \\
\hline
Error (\%) & 32.64 & 25.61 & 18.76 & 37.43 & 30.92 & 26.95 \\
\hline
Standard Deviation & 4.36 & 2.06 & 3.32 & 3.10 & 5.30 & 2.75 \\
\hline
\end{tabular}
\caption{Errors (\%) of generated samples.}
\label{tab:generator_accuracy_svhn_celeba}
\vspace{-1.5em}
\end{table}

On the CelebA data set we also model $u$ as a 128-dimensional vector. Since the CelebA data set has no class labels as such, we choose to encode the facial attributes hair color, gender, smiling and pale skin within $c$. It therefore consists of a five-dimensional categorical variable $c_{cat_1}$ encoding hair color (bald, black, blond, brown, gray), and three two-dimensional categorical variables for the other three attributes. Additionally, we use four continuous variables to encode other (unknown) characteristics. We make sure that our labeled subset of training data is roughly balanced according to the individual labels. This means that approximately one fifth of the labeled samples are sampled from each of the five hair colors, while also ensuring that the samples are split somewhat evenly for the other attributes. The one exception to this rule is the class of \textit{bald} faces, since they are all male.

Fig.~\ref{fig:generator_encoder_acc} shows how the accuracy of the generator develops for different amounts of labels averaged over 10 independent runs and detailed results can be found in Table~\ref{tab:generator_accuracy_svhn_celeba}. The graph only shows the accuracy for generated images according to the characteristics \textit{black, blond, brown hair; smiling, not smiling; male, female}. This is because we found that the generator often fails to generate the characteristics \textit{bald} and \textit{gray hair} correctly. Additionally, images labeled with \textit{pale skin} in the CelebA data set tend to be ``very'' pale. While our generator generates images with pale skin, it was usually not ``pale enough'' for our classifier to classify correctly and it was therefore also left out, but can be qualitatively seen in the generated images. Due to this, we also do not depict the accuracy of the encoder on the training data, since the labels on the CelebA data set are not always accurate and sometimes contradict each other (e.g.\ there are also images that are labeled with multiple hair colors). Still, for the mentioned characteristics we achieve a good accuracy with only 2000 samples, whereas Spurr et al. \cite{spurr2017ecml} report the need for more than 30000 images to achieve a stable training performance.

\begin{figure}[ht]
\footnotesize
\centering
\def\svgwidth{\columnwidth}
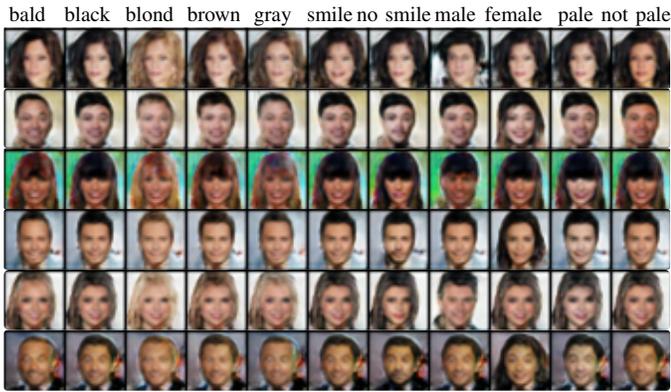
\caption{Image generation (model trained with 1000 labels): in each row the latent representation $u$ is kept constant and only the categorical variables are changed across the columns.}
\label{fig:celeba_random_samples}
\end{figure}

\begin{figure}[ht]
\footnotesize
\centering
\def\svgwidth{\columnwidth}
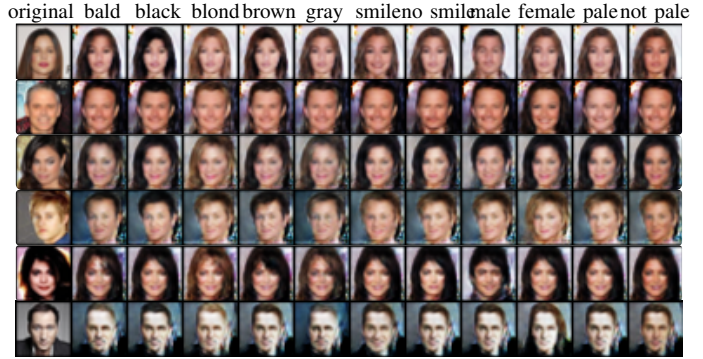
\caption{Image translation: the first column contains randomly sampled images from the CelebA test set while the other columns are translated images where individual categorical variables are changed across columns.}
\label{fig:celeba_class_translation}
\end{figure}

Fig.~\ref{fig:celeba_random_samples} shows images generated according to different settings of $c$, while Fig.~\ref{fig:celeba_class_translation} shows image translations according to the same characteristics. We can see that the characteristic \textit{bald} does not work very well (especially for women). The reason for this is most likely that the labeling on the CelebA data set is not always correct and many images that are labeled as \textit{bald} are not actually bald (more than 25\% of the images labeled as \textit{bald} are also labeled as \textit{gray hair} and there are no bald women in the CelebA data set). We also observe that the attribute \textit{gray hair} often leads the generator to increase the depicted person's age, since there is a high correlation between \textit{gray hair} and age in the CelebA data set (only 380 of 8499 images with the label \textit{gray hair} are also labeled as \textit{young}).

\begin{figure}[ht]
\footnotesize
\centering
\def\svgwidth{\columnwidth}
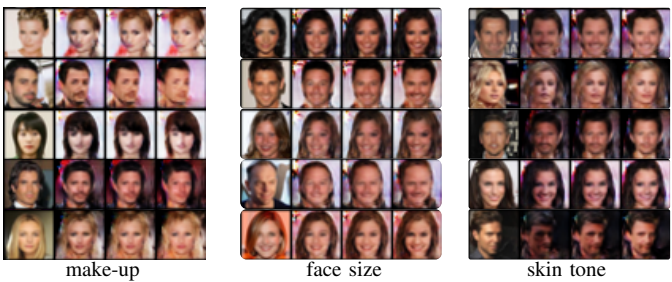
\caption{Image translation: the first column of each block contains randomly sampled images from the CelebA test set. The other columns are translated images where only one of the continuous variables is changed from -1 to 1.}
\label{fig:celeba_translations_cont}
\vspace{-1em}
\end{figure}

On the other hand, the characteristic \textit{blond hair} works better for women than for men. Again, this is most likely due to the fact that more women than men in the CelebA data set are blond (28234 vs.\ 1749) and blond hair tends to be more ``pronounced'' for women. Other characteristics such as \textit{smiling} and gender are modeled very accurately. Finally, Fig.~\ref{fig:celeba_translations_cont} shows translations according to some of the continuous variables which were trained completely without labels. We can see that they learned to encode information such as the amount of applied make-up, the size of the face, and skin tone.

Finally, the image translations do not always translate the face identity correctly, i.e.\ we can see that the facial features change slightly between the original image and the translated ones. This is most likely due to the fact that our reconstruction loss is only a small part of the overall loss and the adversarial loss is usually the dominating factor. As a result, high level features such as e.g.\ facial orientation are reconstructed correctly, while more detailed features are sometimes lost. We find that we can increase the fidelity of the translations by increasing the weight $\lambda_2$ of the reconstruction loss, however, this leads to a slight drop in the accuracy of the generator. At the moment there therefore exists a trade-off between the fidelity of the translations and the accuracy of the generator.

\section{Conclusion}
We developed a system that is capable of both image-to-image translation and controllable image generation. We make use of an encoder which encodes existing images into a latent representation and a generator which takes as input a latent representation and generates an image from it. The latent representation is split up into two parts encoding unstructured information and structured information such as class labels. The structured information is encoded in a disentangled manner and by maximizing the mutual information between this disentangled representation and the images generated from it, we can detect unknown data-generating factors. By specifying the structured information, the image generation process can be controlled, making it useful for generating new images and translating images by adapting the latent representation obtained through the encoder. A discriminator taking as input both a representation and the corresponding image improves the quality of the generated images and encourages the encoder and the decoder to learn inverse function of each other.

Compared to other state-of-the-art image generation and translation systems we can do both, controllable image generation and image-to-image translations with one model. We can translate between multiple domains without the need for multiple encoder-decoder pairs and can even detect novel data-generating factors, which can then be used as additional information for the image generation and translations tasks. For all this we only need a small amount of labeled training samples and we can also make use of only partially labeled data. We test the system on the MNIST, SVHN, and CelebA data sets and show that it is capable of both image translation and controllable image generation across these different data sets. We also find that the system does learn unlabeled data-generating factors such as digit size and rotation or image contrast, which enables us to generate and translate images where we can specify these characteristics, even without any labels provided for them in the training data.

\bibliography{IEEEabrv,./references/references.bib}

\end{document}